\documentclass{article}

\usepackage{arxiv}
\usepackage[utf8]{inputenc}
\usepackage[T1]{fontenc}
\usepackage{hyperref}
\usepackage{url}
\usepackage{booktabs}
\usepackage{amsfonts}
\usepackage{nicefrac}
\usepackage{microtype}
\usepackage{graphicx}
\usepackage[numbers,sort&compress]{natbib}
\graphicspath{{./figs/}}

\usepackage{amsmath,amsfonts,bm}

\def\eqref#1{equation~\ref{#1}}

\def\1{\bm{1}}

\DeclareMathAlphabet{\mathsfit}{\encodingdefault}{\sfdefault}{m}{sl}
\SetMathAlphabet{\mathsfit}{bold}{\encodingdefault}{\sfdefault}{bx}{n}

\title{BiasMix-Finance: Post-Generation KYC Guardrails for LLM Portfolio Advice}
\date{}
\renewcommand{\shorttitle}{BiasMix-Finance: Post-Generation KYC Guardrails for LLM Portfolio Advice}

\author{Gaurav Kukreja \\
\texttt{gauravkukreja06@yahoo.in} \\
\And
Parul Kukreja \\
\texttt{parulsehgal11@gmail.com} \\
\AND
Mohammed Abraar \\
\texttt{abraar@vizz.vizuara.ai} \\
\AND
Raj Dandekar \\
\texttt{raj@vizuara.com} \\
\AND
Rajat Dandekar \\
\texttt{rajatdandekar@vizuara.com} \\
\AND
Sreedath Panat \\
\texttt{sreedath@vizuara.com} \\
}

\begin{document}
\maketitle
\begin{abstract}
Large language models (LLMs) can generate plausible-sounding ETF portfolios while silently violating basic KYC-style constraints on risk, fees, and diversification. This is especially problematic in \emph{agentic} multi-turn advisory systems, where each draft recommendation can become an action unless guarded by an auditable enforcement layer.
We study a model-agnostic, asset-agnostic \emph{post-generation guardrail} pipeline: (i) enforce a strict JSON allocation schema, (ii) validate allocations against numeric caps, and (iii) when violations occur, \emph{deterministically} project the output to the nearest feasible portfolio via a convex quadratic program (QCQP). We introduce \textbf{BiasMix-Finance (Mini)}, a compact stress-test benchmark for \emph{constrained decision-making under biased LLM generations}, with a 16-ETF universe, three investor profiles, and eight bias prompts.
Across three models and three inference modes (direct, critique, self-consistency), first-pass generations violate at least one cap in \textbf{47.6--85.7\%} of test cases (\textbf{67.2\%} pooled), but the convex projection layer reduces \emph{final feasibility violations to \textbf{0\%}} while requiring only a \emph{small correction distance} (test pooled median $D=\|w^* - w_0\|_2=0.066$), indicating that the guardrail typically preserves the intent of the original allocation.
We report violation rates and correction distances with confidence intervals, and paired model comparisons with multiple-testing correction. To support reproducibility, we release the dataset, prompts, caps, and code in our public GitHub \href{https://github.com/gauravkukreja06/biasmix-finance}{repository}.
\end{abstract}

\section{Introduction}
Large language models (LLMs) are increasingly used to generate financial summaries, recommendations, and portfolio allocations \citep{wu2023bloomberggpt,yang2023fingpt}.
However, in regulated advisory settings, recommendations are expected to reflect customer-specific facts and suitability obligations \citep{finra2090,finra2111}. In our controlled setting, we operationalize this idea as hard numeric caps on risk, fees, and diversification, which are also standard in constrained portfolio construction \citep{markowitz1952,jagannathan2003nogo,boyd2004convex}.
Prompting and reasoning-mode variations (e.g., critique, self-consistency) are inherently probabilistic and cannot guarantee constraint satisfaction on every run \citep{wei2022cot,wang2022selfconsistency,madaan2023selfrefine}.
This motivates \textbf{post-generation enforcement}: we treat the LLM output as a draft allocation and apply deterministic verification and repair, combining audit-style validation with convex feasibility projection \citep{raji2020closing,ribeiro2020checklist,boyd2004convex}.
Crucially, our goal is \emph{not} to solve a classical portfolio-optimization problem (e.g., maximizing risk-adjusted return under constraints) \citep{markowitz1952,blacklitterman1992global,goldfarb2003robust}.
Instead, we study an \emph{auditing and repair} problem: how to reliably detect and minimally correct \emph{irrational or constraint-violating} allocations produced by an LLM, while preserving the model's intended allocation structure as much as possible \citep{ribeiro2020checklist,bender2021stochastic,raji2020closing}.
More broadly, this setting is adjacent to LLM alignment and policy-following work \citep{ouyang2022instruct,bai2022constitutional}, but differs because correctness depends on satisfying externally specified constraints that can be verified by deterministic checks.
Because these constraints are externally specified and mechanically checkable, the task is a natural testbed for \emph{post-hoc verification and repair} pipelines that are model-agnostic and transferable to other constrained decision-making domains (e.g., resource allocation, scheduling, or policy compliance).
We study a controlled portfolio-allocation setting: given an ETF universe and a scenario describing (i) a risk profile with caps and (ii) a bias-inducing context (e.g., ``small-cap hype''), the LLM emits portfolio weights under a strict JSON schema.
A key challenge is that biased or preference-shaping contexts can systematically push generations toward concentrated, high-volatility, or otherwise non-compliant allocations.
To evaluate guardrails under these realistic failure modes, we adopt a \textbf{bias-induction methodology}: we programmatically generate scenarios that combine standardized risk profiles with bias ``recipes'' (e.g., anchoring on a sector, FOMO tilts, inertia, fee neglect), producing a stress-test suite for constraint-violating drafts.
We then measure constraint violations \emph{before} repair and the magnitude of correction required \emph{after} repair.
Our goal is to quantify how much prompt reasoning modes reduce violations \emph{prior} to enforcement, and how much post-hoc guardrails must change the portfolio to make it compliant.
In this sense, we treat BiasMix-Finance (Mini) as a stress-test benchmark for \emph{constrained decision-making under biased generations}, complementing broader evaluation efforts in safe and reliable LLM deployment. To verify that BiasMix prompts elicit the intended behavioral tilts, Appendix Table~\ref{tab:bias-examples} shows representative first-pass LLM outputs (held-out test) for each bias recipe. 
We focus on post-generation enforcement, which is especially relevant for \emph{agentic} financial advisors that operate in multi-turn loops and tool-call to validate actions.
Section~\ref{sec:agentic} discusses how deterministic verify-and-repair acts as a governance mechanism inside autonomous advisor systems.

\paragraph{Contributions.}
We emphasize that the contribution is a \emph{methodology and evaluation framework} for deploying LLMs in constrained financial decision workflows, rather than a new optimization technique, the QCQP projection is used as a deterministic repair operator.
\begin{itemize}
  \item \textbf{Treating LLM outputs as auditable drafts:}
  We treat LLM-generated allocations as \emph{auditable drafts} that must satisfy explicitly specified, mechanically checkable KYC-style numeric caps, enabling transparent logging of (draft, violations, repaired output) for governance and oversight.

  \item \textbf{Bias-induced stress testing as an evaluation primitive:}
  We construct 72 scenario instances (train/dev/test) that combine three risk profiles with eight behavioral ``bias recipes'' and three random seeds, over a fixed 16-ETF universe.
  This controlled design isolates how different prompts and models respond to the same constraints and bias contexts, and makes bias induction itself a reproducible evaluation primitive for constrained decision-making under LLM outputs.

  \item \textbf{Deterministic repair as a first-class safety mechanism:}
  We formalize the feasible set induced by volatility, fee, HHI concentration, and max single-asset/sector caps, and compute the nearest feasible portfolio via a constrained quadratic program, solved with standard convex optimization tools \citep{boyd2004convex,diamond2016cvxpy,odonoghue2016conic,stellato2020osqp}.

  \item \textbf{Reproducible statistical evaluation across splits, models, and modes:}
  We report per-cap violation rates with Wilson 95\% confidence intervals \citep{wilson1927}, correction-distance and metric-delta confidence intervals via bootstrap \citep{efron1993}, and paired model comparisons on correction distance with Wilcoxon tests plus BH-FDR correction \citep{wilcoxon1945,benjamini1995}.
\end{itemize}

\section{Problem Setup}
\label{sec:problem_setup}
\textbf{Universe.} Let $n$ denote the number of ETFs in the universe (here $n=16$) and let $w \in \mathbb{R}^n$ denote portfolio weights.

\textbf{Scenario.} Each scenario provides (i) a risk profile with caps and (ii) a bias context string intended to nudge the model toward potentially non-compliant allocations.
Caps include an annualized volatility ceiling $\sigma_{\mathrm{cap}}$, a weighted-average expense ratio ceiling $f_{\mathrm{cap}}$, a concentration ceiling $h_{\mathrm{cap}}$ based on the Herfindahl--Hirschman Index (HHI) \citep{herfindahl1950,hirschman1964}, and max single-asset and max single-sector weight caps. Such constraints are standard in constrained portfolio construction and are commonly studied as practical overlays on mean--variance style allocations \citep{markowitz1952,jagannathan2003nogo}.

\textbf{Model output.} The LLM emits a draft portfolio $w_0$ (JSON with nonnegative weights that sum to 1).
We compute summary metrics: risk $\sigma(w)=\sqrt{w^\top \Sigma w}$ \citep{markowitz1952}, fee $\mathrm{WAER}(w)=\sum_i w_i \cdot \mathrm{fee}_i$, and concentration $\mathrm{HHI}(w)=\sum_i w_i^2$.

\textbf{Feasible set.} Let $\mathcal{C}$ be the set of portfolios satisfying all caps.
A draft is \emph{violating} if $w_0 \notin \mathcal{C}$.

\textbf{Post-generation enforcement.} When $w_0$ violates any cap, we compute the nearest feasible portfolio
\begin{equation}
w^* = \arg\min_{w} \lVert w - w_0 \rVert_2^2 \quad \text{s.t.} \quad w \in \mathcal{C},
\end{equation}
and define the correction distance as $D=\lVert w^* - w_0 \rVert_2$ (Euclidean/L2 distance). Projection operators are a standard way to enforce feasibility under convex constraints \citep{boyd2004convex,duchi2008efficient}, and related portfolio work often uses robust or constrained formulations to stabilize allocations under estimation error \citep{blacklitterman1992global,goldfarb2003robust,bental2009robust}.

\section{Related Work}
\paragraph{LLMs for finance.}
Domain-specific financial LLMs such as BloombergGPT \citep{wu2023bloomberggpt} and open efforts like FinGPT \citep{yang2023fingpt} show strong capability on financial NLP tasks; our focus differs in that we study \emph{hard numeric constraint compliance} for portfolio-weight outputs rather than text generation accuracy.

\paragraph{Guardrails, auditing, and constrained outputs.}
Safety toolkits and classifiers (e.g., NeMo Guardrails, Llama Guard) \citep{rebedea2023nemoguardrails,meta2023llamaguard} and alignment methods such as RLHF and Constitutional AI \citep{ouyang2022instruct,bai2022constitutional} primarily target content/policy enforcement.
Complementary work emphasizes \emph{auditing} and \emph{verification} of LLM outputs in high-stakes settings via structured outputs, validation checks, and feedback loops \citep{raji2020closing,ribeiro2020checklist}, and constrained/guided decoding provides guarantees for lexical or structural constraints \citep{hokamp2017gridbeam,willard2023efficientguided}.
Our setting requires satisfying quantitative constraints naturally expressed as convex restrictions over portfolio weights; this motivates optimization-based post-processing as an auditable repair layer \citep{boyd2004convex}.
We discuss additional related work and context in Appendix~\ref{app:relwork}.

\section{Methodology}
\label{sec:setup}
\begin{figure}[t]
\centering
\includegraphics[width=\linewidth]{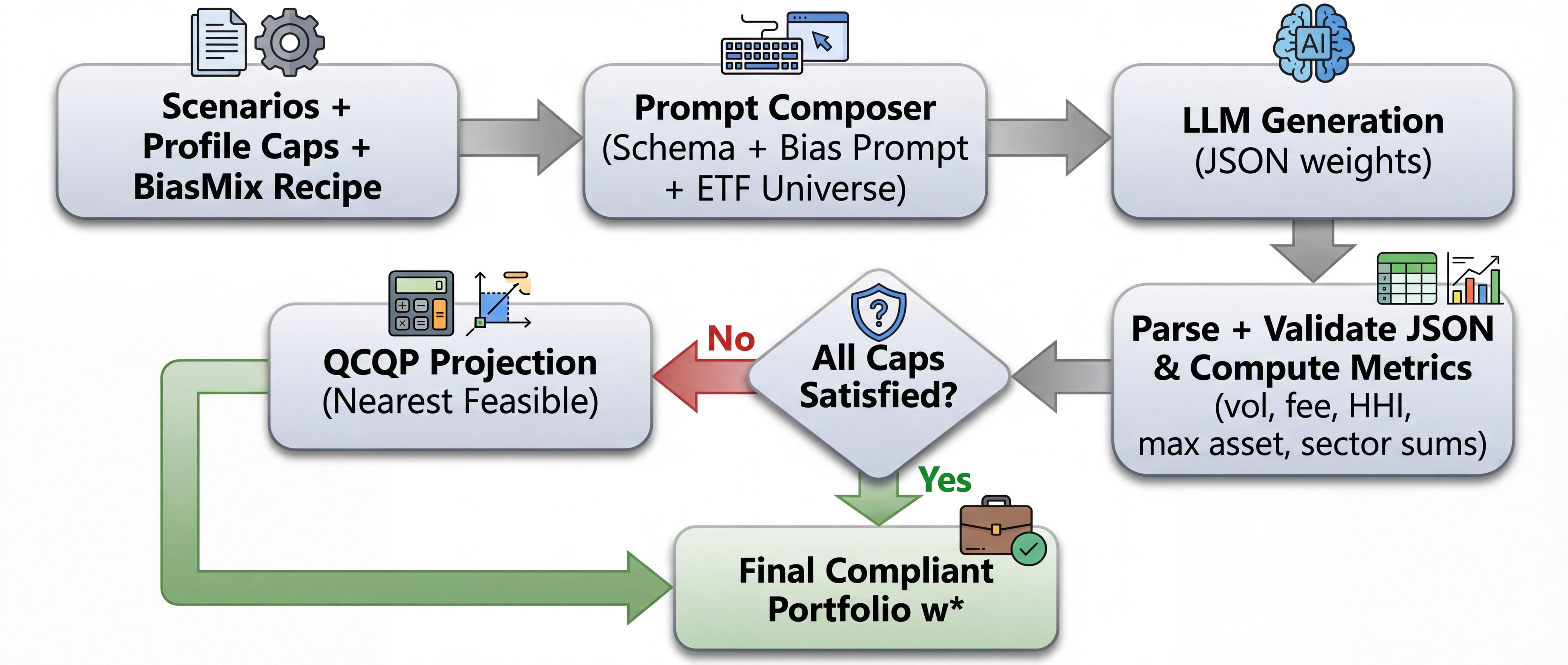}
\caption{\textbf{End-to-end pipeline.} BiasMix scenario + bias recipe are composed into a structured prompt. The LLM generates draft weights (JSON). We validate JSON and compute portfolio metrics against hard KYC caps. If any constraint fails, we repair via convex projection (QCQP) to the nearest feasible portfolio; otherwise we accept directly.}
\label{fig:pipeline}
\end{figure}

Figure~\ref{fig:pipeline} summarizes the end-to-end BiasMix-Finance pipeline from scenario construction to deterministic feasibility via projection \citep{boyd2004convex,ribeiro2020checklist}.

\subsection{Universe and scenarios}
We construct a synthetic-but-controlled scenario suite over a fixed ETF universe of $n=16$ liquid funds:
\texttt{SPY, VEA, VWO, VGT, XLE, XLF, XLV, XLY, XLP, XLI, XLRE, IWM, AGG, LQD, IEF, GLD}.
Each scenario specifies an \emph{as-of} date (we use \texttt{2025-09-30}), a risk profile
(Conservative/Moderate/Aggressive), and a \emph{BiasMix} recipe that nudges the model toward a known behavioral
bias (e.g., ``anchor tech'', ``FOMO energy'', ``small-cap hype''). Recipes are expressed as short natural-language
preferences (not hard constraints) so that first-pass outputs can be either feasible or infeasible. This design parallels behavioral stress testing in NLP evaluation: the goal is to systematically elicit failure modes under controlled perturbations \citep{ribeiro2020checklist}. We generate 72 scenarios across 3 profiles and 8 bias recipes, with \emph{three} random seeds per (profile, recipe) cell; split sizes are train/dev/test = 36/15/21, balanced across the three risk profiles (each profile contributes 12/5/7 to train/dev/test). We use train only for prompt and hyperparameter iteration (e.g., mode temperatures, self-consistency sample size K, and projection margins). Dev a one-time check that the tuned pipeline behaves similarly on unseen data. We report results on test without further tuning. Table~\ref{tab:scenario_breakdown} summarizes the dataset. We use train/dev/test splits only to avoid leakage from repeated prompt/solver tuning; split construction details are in Appendix~B.5.

\begin{table}[t]
\caption{BiasMix scenario breakdown (72 total) by risk profile and bias type. Each (profile, bias) cell has 3 scenarios (3 random seeds).}
\label{tab:scenario_breakdown}
\vspace{\baselineskip}

\begin{center}
\small
\setlength{\tabcolsep}{3pt}
\begin{tabular}{lccccccccc}
\multicolumn{1}{c}{\bf Profile} & \bf Anchor & \bf Default & \bf EM & \bf FOMO & \bf Fee & \bf Gold & \bf Small--cap & \bf US--only & \bf Total \\
\multicolumn{1}{c}{} & \bf tech & \bf inertia & \bf tilt & \bf energy & \bf neglect & \bf craze & \bf hype & \bf bias & \multicolumn{1}{c}{} \\
\hline
Conservative & 3 & 3 & 3 & 3 & 3 & 3 & 3 & 3 & 24 \\
Moderate     & 3 & 3 & 3 & 3 & 3 & 3 & 3 & 3 & 24 \\
Aggressive   & 3 & 3 & 3 & 3 & 3 & 3 & 3 & 3 & 24 \\
Total        & 9 & 9 & 9 & 9 & 9 & 9 & 9 & 9 & 72 \\
\end{tabular}
\end{center}
\end{table}

\subsection{Risk profiles and hard caps}
All caps are defined over the portfolio functions in Section~\ref{sec:problem_setup}. For each risk profile we fix
a cap vector \(\theta=(\sigma_{\mathrm{cap}}, f_{\mathrm{cap}}, h_{\mathrm{cap}}, a_{\max}, s_{\max})\)
covering annualized volatility, weighted-average expense ratio (WAER), concentration (HHI), and single-asset / single-sector limits.
Table~\ref{tab:caps} lists the numerical values used in all experiments.

\noindent\textbf{Cap rationale}
The caps in Table~\ref{tab:caps} are chosen as representative KYC-style suitability heuristics inspired by customer-specific suitability and diversification principles, not jurisdiction-specific rules \citep{finra2090,finra2111,jagannathan2003nogo}; detailed justification and threshold sensitivity are in Appendix~\ref{app:caps}.

\begin{table}[t]
\caption{Risk-profile caps used throughout. $\sigma_{\mathrm{cap}}$ is annualized volatility, WAER is weighted-average expense ratio,
HHI is concentration, $a_{\max}$ is max single-asset weight, and $s_{\max}$ is max single-sector weight.}
\label{tab:caps}
\vspace{\baselineskip}
\begin{center}
\small
\setlength{\tabcolsep}{6pt}
\begin{tabular}{lccccc}
\multicolumn{1}{c}{\bf Profile} & \multicolumn{1}{c}{\bf $\sigma_{\mathrm{cap}}$} & \multicolumn{1}{c}{\bf $f_{\mathrm{cap}}$ (WAER)} & \multicolumn{1}{c}{\bf $h_{\mathrm{cap}}$ (HHI)} & \multicolumn{1}{c}{\bf $a_{\max}$} & \multicolumn{1}{c}{\bf $s_{\max}$}
\\ \hline
Conservative & 0.06 & 0.0020 & 0.12 & 0.25 & 0.30 \\
Moderate     & 0.10 & 0.0030 & 0.18 & 0.35 & 0.40 \\
Aggressive   & 0.14 & 0.0040 & 0.25 & 0.45 & 0.50 \\
\end{tabular}
\end{center}
\end{table}

\subsection{Guardrail pipeline}
Given a scenario, the model outputs portfolio weights \(w_0\) under a strict JSON schema.
We validate \(w_0\); if any constraint in \(\mathcal{C}(\theta)\) is violated, we compute a nearest-feasible projection:
\begin{equation}
w^* \;=\; \arg\min_{w\in\mathcal{C}(\theta)} \; \lVert w - w_0 \rVert_2^2 \;+\; \lambda\, w^\top \Sigma w,
\label{eq:projection}
\end{equation}
where feasible set $\mathcal{C}$ is induced by the hard caps:
\begin{equation}
\mathcal{C} = \left\{ 
w \in \mathbb{R}^n : 
\begin{aligned}
    & \mathbf{1}^\top w=1, \; w\ge 0, \\
    & \sigma(w)\le\sigma_{\text{cap}}, \; \mathrm{WAER}(w)\le f_{\text{cap}}, \\
    & \mathrm{HHI}(w)\le h_{\text{cap}}, \; \max\nolimits_i w_i \le a_{\text{cap}}, \\
    & \text{sector\_sum}(w)\le s_{\text{cap}}
\end{aligned}
\right\}
\end{equation} 
\(\lambda=2\times 10^{-3}\) is a small variance regularizer that discourages solutions sitting exactly on the risk boundary.
We report the \emph{correction distance} \(D=\lVert w^* - w_0\rVert_2\) (Euclidean norm).
We solve the projection as a convex QCQP with standard solvers; numerical stability choices (PSD jitter), solver tolerances, fallback strategy, and robustness ablations are reported in the Appendix~\ref{app:solver}

\subsection{Models and providers}
We evaluate three LLM backends: (i) \emph{Gemini 2.5 Flash} (\texttt{gemini-2.5-flash}) via the Google GenAI Python SDK,
(ii) \emph{GPT-5 nano} (\texttt{gpt-5-nano}) via the OpenAI Chat Completions API, and
(iii) an open-weight model served through an OpenAI-compatible endpoint, \emph{Llama 3.3 70B Instruct Turbo}
(\texttt{meta-llama/Llama-3.3-70B-Instruct-Turbo}).

\subsection{Inference Strategies (Modes)}
The unit of analysis is a \emph{scenario} evaluated under a (model, mode) configuration. We evaluate the models across three strategies: 
(i) \textbf{Direct} generation, where the model is prompted to produce the target JSON output in a single shot; 
(ii) \textbf{Critique}, a multi-turn approach where the model generates an internal draft and critique before emitting a corrected JSON \citep{madaan2023selfrefine}; and 
(iii) \textbf{Self-consistency (SC)}, where $K=5$ candidates are sampled, and the candidate with the lowest pre-projection violation is selected \citep{wang2022selfconsistency}.
All modes share the same post-generation projection step when constraints are violated and enforce strict JSON outputs with up to R=3 retries on parse failure; once parsed, weights are validated and projected if needed. Full decoding and parsing details are in the Appendix~\ref{app:reproducibility}.

\section{Experimental protocol and statistical framework}
\label{sec:experiments}
We evaluate each scenario under a (model, mode) configuration and measure compliance on the first-pass draft $w_0$ and after deterministic repair $w^*$.
Our \textbf{primary endpoints} are (i) first-pass violation rate (per-cap and any-cap) and (ii) correction distance $D=\lVert w^*-w_0\rVert_2$.
We additionally report parse-failure rate and before/after deltas in $\sigma$, WAER, and HHI.
For violation rates we use Wilson 95\% CIs \citep{wilson1927}; for $D$ and deltas we use bootstrap 95\% CIs \citep{efron1993}.
For paired model comparisons on $D$ we use Wilcoxon signed-rank tests with BH-FDR correction \citep{wilcoxon1945,benjamini1995,demsar2006}.
Full protocol details (splits, resampling, and multiple-testing family) are in Appendix~\ref{app:statistical-details}.

\section{Results}
\subsection{First-pass violations on held-out test}
\begin{figure}[t]
\begin{center}
    \includegraphics[width=0.95\linewidth]{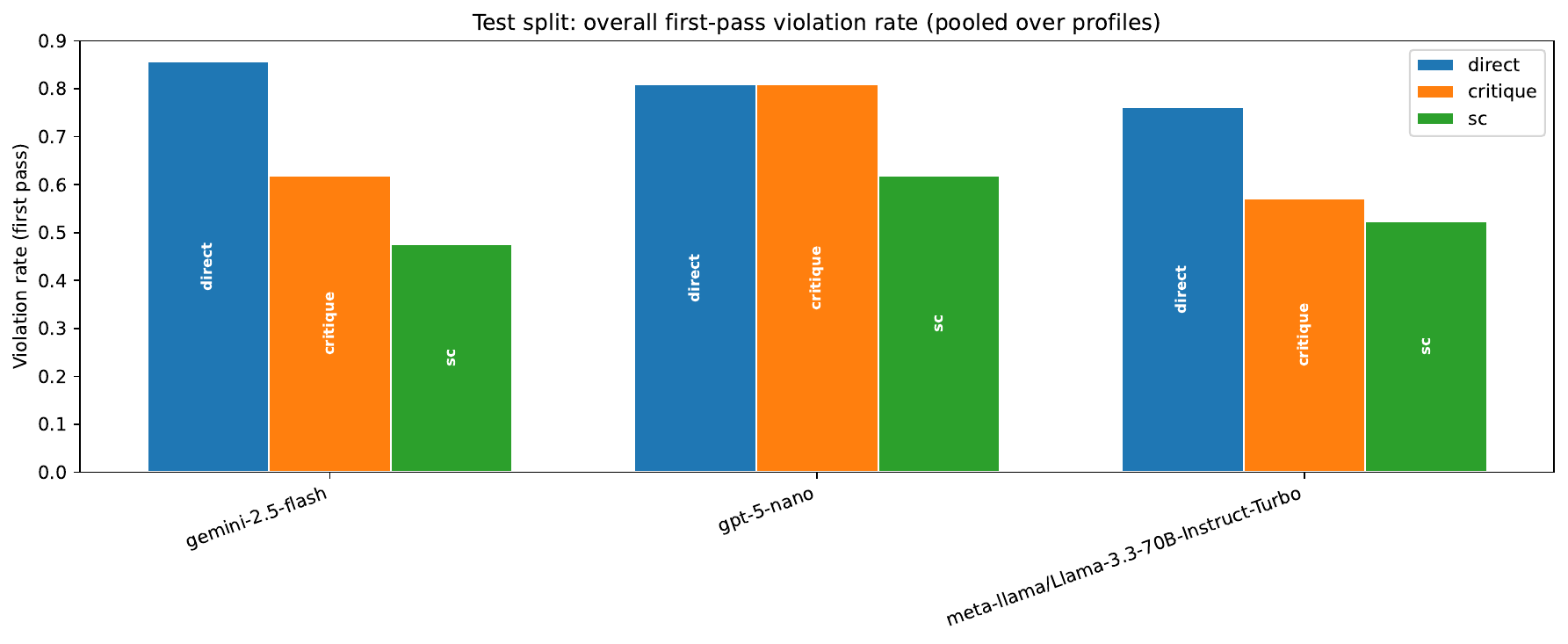}
\end{center}
\caption{Held-out test: overall first-pass violation rate (pooled over profiles) with Wilson 95\% CIs.}
\label{fig:viol}
\end{figure}

Table~\ref{tab:test-viol} and Figure~\ref{fig:viol} report pooled first-pass violation rates on the test split. Critique and self-consistency tend to reduce violations relative to direct prompting, but do not guarantee compliance. Violation rates are heterogeneous across BiasMix recipes and profiles; Appendix Figs.~\ref{fig:bias-heatmap-direct}--\ref{fig:bias-heatmap-sc} visualize test-split violation heatmaps by bias type, profile, and model (for each prompting mode).
Interpreting the magnitudes, prompting alone remains unreliable under hard KYC-style caps: even the best test setting (Gemini+SC) violates in nearly half of cases ($0.476$), while the worst (Gemini+direct) violates most of the time ($0.857$).
Self-consistency helps most when the feasible region is larger (Aggressive drops from $0.476$ to $0.048$), but tight regimes remain difficult: Conservative portfolios violate on all test scenarios across models/modes ($1.0$), motivating a deterministic verify-and-repair layer. The Conservative regime is intentionally \emph{extremely tight}: the intersection of a low volatility cap, a strict HHI concentration cap, and sector/asset limits induces a feasible region that is very small on the simplex.
In practice, this means the admissible set resembles a narrow polytope/conic slice that occupies a tiny fraction of the space of plausible-looking portfolios.
Because prompting and reasoning modes remain probabilistic, they cannot reliably ``hit'' such a small feasible region in one shot, even when the generated portfolio appears sensible at a glance.
This is precisely the setting where deterministic post-generation repair is necessary: it converts an intent-bearing but infeasible draft into a compliant action while minimizing deviation from the draft, making hard-cap satisfaction a property of the system interface rather than the sampling behavior of the model.

\subsection{Projection yields high final feasibility}
Most failures arise at parse time (invalid JSON), not at constraint time.
On the held-out test split, with strict schema prompting and up to $R=3$ retries, we observe
\emph{zero parse failures} and \emph{100\% post-projection feasibility} across all models and modes
(Wilson 95\% CIs: parse-fail $[0,0.155]$, final pass $[0.845,1.0]$; full table in Appendix~\ref{app:robustness}).
Thus, reliability hinges primarily on structured output; once weights are parsed, solver-based repair
enforces constraints deterministically while preserving the LLM as an intent generator.

\subsection{How much correction is required?}
Table~\ref{tab:test-D} quantify the adjustment required to reach feasibility, the corresponding visualization is in Appendix~\ref{app:correction-distance}.
Lower $D$ indicates the model proposed a near-feasible portfolio.
We interpret $D$ as a faithfulness signal: small $D$ means projection only nudges the draft onto the feasible set, while large $D$ means constraints substantially override the draft.
Consistent with tightness, corrections are smallest for Aggressive (often near zero, especially with SC), moderate for Moderate (roughly $0.018$--$0.185$), and largest for Conservative (roughly $0.220$--$0.462$), aligning with its near-certain first-pass violations.
In deployment, large $D$ can be surfaced as ``constraint-forced rebalancing,'' while small $D$ supports the claim that guardrails can guarantee compliance without materially changing near-feasible drafts. Beyond L2 distance, intent is also preserved at a coarser granularity: Appendix Figs.~\ref{fig:sector-preserve-direct}--\ref{fig:sector-preserve-sc} plot draft vs.\ projected \emph{sector} allocations on the test split, with points concentrated near the diagonal.
On the held-out test split (pairs=21), differences are significant in \textit{direct} and \textit{critique} but not \textit{SC}:
gpt-5-nano vs meta-llama/Llama-3.3-70B-Instruct-Turbo is significant in direct ($p_{\mathrm{FDR}}{=}0.0292$) and critique ($p_{\mathrm{FDR}}{=}0.0098$), and Gemini-2.5-flash vs meta-llama/Llama-3.3-70B-Instruct-Turbo is also significant in direct ($p_{\mathrm{FDR}}{=}0.0669$) and critique ($p_{\mathrm{FDR}}{=}0.0098$).
Full results across splits and modes are in Appendix~\ref{tab:wilcoxon_all}.

\begin{table}[t]
\caption{\textbf{Test correction distance $D$ (bootstrap 95\% CI).} $D=\lVert w^*-w_0\rVert_2$ (unitless L2 distance on the simplex); lower values mean the model output was closer to feasible before projection.}
\label{tab:test-D}
\vspace{\baselineskip}
\begin{center}
\small
\setlength{\tabcolsep}{4pt}
\begin{tabular}{@{}lllll@{}}
\multicolumn{1}{c}{\bf Model} & \multicolumn{1}{c}{\bf Mode} & \multicolumn{1}{c}{\bf $n$} & \multicolumn{1}{c}{\bf $D$ mean (band)} & \multicolumn{1}{c}{\bf $D$ median}\\
\hline
gemini-2.5-flash & critique & 21 & 0.108 [0.000,0.383] & 0.028 \\
gemini-2.5-flash & direct & 21 & 0.186 [0.009,0.591] & 0.048 \\
gemini-2.5-flash & sc & 21 & 0.104 [0.000,0.376] & 0.000 \\
gpt-5-nano & critique & 21 & 0.127 [0.007,0.320] & 0.121 \\
gpt-5-nano & direct & 21 & 0.203 [0.012,0.435] & 0.110 \\
gpt-5-nano & sc & 21 & 0.097 [0.000,0.276] & 0.072 \\
Llama-3.3-70B-Instruct & critique & 21 & 0.106 [0.000,0.331] & 0.029 \\
Llama-3.3-70B-Instruct & direct & 21 & 0.136 [0.000,0.358] & 0.119 \\
Llama-3.3-70B-Instruct & sc & 21 & 0.097 [0.000,0.322] & 0.028 \\
\end{tabular}
\end{center}
\end{table}

\section{Implications for Agentic Financial AI}
\label{sec:agentic}

FinAI emphasizes \emph{agentic} financial systems: autonomous or semi-autonomous advisors that interact over multiple turns, call tools, and iteratively refine decisions under constraints.
Although BiasMix-Finance is evaluated as a post-generation guardrail for a single draft, the same verify-and-repair interface naturally composes with agent loops.
In this view, the LLM is an \emph{intent generator} (proposal policy) and the guardrail is an \emph{enforcement tool} that ensures every externally visible action satisfies hard KYC-style constraints.

\paragraph{Multi-turn advisor interactions.}
As users update preferences or facts (e.g., fee sensitivity, horizon, concentration tolerance), a multi-turn agent can treat each intermediate allocation as a tentative action, run validation and projection, and return a feasible portfolio at \emph{every turn}.
Even if intermediate reasoning is noisy, deployed outputs remain within a clearly specified feasible region.
The correction distance $D=\|w^*-w_0\|_2$ can be surfaced as an explanation signal for how strongly constraints overruled the draft (``light nudge'' vs.\ ``constraint-forced rebalancing'').

\paragraph{Iterative correction loops in autonomous agents.}
Agentic systems often operate as repeated propose--check--revise cycles \citep{yao2023react,madaan2023selfrefine}.
Our pipeline provides a deterministic inner loop: propose an allocation, validate, project to feasibility, and feed the corrected portfolio back into the agent state.
This replaces brittle prompt-only self-correction with a tool-grounded mechanism that enforces admissible actions whenever projection succeeds.

\paragraph{Guardrails as governance mechanisms.}
In regulated settings, governance requires policies that are auditable, stable, and configurable (e.g., per institution, jurisdiction, or product type).
Numeric caps and deterministic repair provide a concrete governance layer: constraints are explicit and versionable, and every recommendation can be logged with (draft, violations, repair, and $D$).

\paragraph{Feedback adaptation from corrected feasible outputs.}
The (draft $\rightarrow$ projected) pairs produced by the guardrail are high-signal supervision: they encode how a recommendation must change to satisfy policy.
This supports (1) \emph{prompt-level adaptation} that conditions on prior corrections to draft closer-to-feasible portfolios (reducing $D$), and (2) \emph{model-level adaptation} using corrected outputs for fine-tuning or preference learning, while retaining repair as a backstop.

\paragraph{Regulatory sandboxing for agent-based financial advisors.}
BiasMix-Finance can serve as a lightweight \emph{regulatory sandbox} by simulating multi-turn dialogues: vary the investor profile, inject biased or adversarial instructions, and measure (i) first-pass violation propensity, (ii) correction distance $D$, and (iii) stability of sector/asset allocations after repair.
Because the enforcement layer is deterministic and policy-driven, the same sandbox can be repeated under different cap configurations to stress-test agent behavior as governance tightens or loosens.

\noindent
Overall, the contribution is not only a reliable post-processor for static drafts, but a composable \emph{agent tool} for financial decision-making: it turns probabilistic proposals into policy-compliant actions for safer multi-turn and autonomous advisory workflows.

\section{Discussion, limitations, and future work}
Prompting modes (direct, critique, self-consistency) can reduce first-pass violations, but they cannot guarantee satisfaction of hard numeric caps because generation remains probabilistic.
In contrast, the post-generation projection step provides an auditable, model-agnostic enforcement layer that deterministically returns a feasible portfolio (when the cap set is feasible), while minimizing deviation from the model’s proposal.

\textbf{Scope and limitations.}
This study is intentionally scoped to \emph{constraint compliance and robustness}, not portfolio optimality: we test whether LLM allocations can be made KYC-feasible under explicit caps and quantify the minimal adjustment required (correction distance $D$).
We therefore do not model expected returns, transaction costs, taxes, or utility-based objectives.
We use a fixed \emph{16-ETF} universe as a controlled stress test; the guardrail is not tied to 16 assets and applies to any universe with fees, sector mappings, and a covariance estimate.
We also do not conduct real-user or advisor-in-the-loop studies, focusing instead on an auditable enforcement mechanism for downstream workflows.

Additional discussion on universe design and generalization is provided in Appendix~\ref{app:scope}.

\paragraph{Transferability beyond portfolio allocation.}
The verify and repair pattern studied here applies to a broader class of \emph{agentic financial decision systems} where an LLM proposes a structured action but correctness depends on hard, externally specified constraints.
Examples include budget allocation under category caps, product eligibility and suitability checks, policy-compliant trade lists (e.g., prohibited assets, minimum lot sizes, concentration and exposure limits), disclosure and fee-rule compliance, and structured credit/insurance recommendation drafts that must satisfy underwriting or regulatory rules.
Across these settings, the key systems principle is to separate \emph{proposal generation} (probabilistic, model-driven) from \emph{admissibility enforcement} (deterministic, auditable).
BiasMix-style stress testing also transfers by inducing known behavioral pressures (e.g., fee neglect, overconfidence, anchoring, recency) and measuring (i) first-pass violation propensity and (ii) the magnitude of repair required to restore compliance.
Thus, the primary contribution is a reusable governance-oriented framework: explicit constraints, deterministic repair, and measurable faithfulness between draft and compliant action.

\textbf{Future work.}
Promising extensions include (i) scaling to larger universes and adding liquidity/turnover constraints, (ii) replacing the fixed covariance with rolling or regime-aware estimates, (iii) incorporating return-aware or utility-based objectives while preserving hard caps, and (iv) running human-evaluation studies to assess whether corrected portfolios preserve user intent and improve trust and usability.

\section{Conclusion}
Post-generation KYC-style guardrails offer a practical and auditable way to enforce numeric investment constraints on top of LLM-generated allocations.
Across models and prompting modes, we find that a convex nearest-feasible projection can eliminate final constraint violations while keeping repairs bounded in L2 distance.
More generally, the same verify-and-repair pattern is \emph{model-agnostic} and \emph{asset-agnostic}: any LLM that emits a structured decision can be checked against hard constraints and deterministically repaired when needed, making the approach applicable beyond finance.
BiasMix-Finance (Mini) serves as a stress-test benchmark for constrained decision-making under bias, enabling controlled comparisons of reasoning modes and models under identical caps and contexts.

\paragraph{LLM usage disclosure.}
We used large language models as auxiliary tools for language polishing and limited code assistance (e.g. code utilities, grammar, and LaTeX formatting). All experimental design, implementation, data collection, analysis, results, and conclusions were produced and verified by the authors, who take full responsibility for the content of this paper.

\bibliography{references}
\bibliographystyle{bibliography_style}

\appendix
\section{Appendix}
\section{Prompts, caps, and scenario template}
\label{app:prompts}

\subsection{Cap rationale and sensitivity}\label{app:caps}
The cap values in Table~\ref{tab:caps} are intended to be representative of practical ``KYC-style'' suitability heuristics rather than jurisdiction-specific regulations \citep{finra2090,finra2111}.
They reflect: (i) lower volatility targets for conservative investors;
(ii) a low-cost fee budget typical of broad-market ETFs (e.g., 0.20--0.40\% WAER);
(iii) diversification via concentration control (HHI corresponds to an effective number of holdings of roughly \(1/\mathrm{HHI}\));
and (iv) concentration limits that prevent over-exposure to a single fund or sector.
(We report threshold sensitivity analyses in this appendix.)

\subsection{Direct/Critique JSON schema}
\begin{quote}
\texttt{\{"weights": \{"TICKER": 0.00, "...": 0.00\}, \\ 
"explanation": "<=120 words"\}}
\end{quote}

\subsection{SC JSON schema}
\begin{center}
\small
\begin{minipage}{0.9\linewidth}
\begin{itemize}
    \item[] \texttt{\{}
    \item[] \quad \texttt{"candidates": [\{"weights": \{"TICKER": 0.00, ...\}, "explanation": "<=120 words"\},}
    \item[] \quad \quad \texttt{...}
    \item[] \quad \texttt{]}
    \item[] \texttt{\}}
\end{itemize}
\end{minipage}
\end{center}

\subsection{Example BiasMix prompts}
We include two representative BiasMix contexts (verbatim) used to induce biased first-pass allocations:
\begin{itemize}
\item \textbf{Anchor tech:} \emph{Strongly prefer VGT (tech tilt).}
\item \textbf{Small-cap hype:} \emph{Strongly prefer IWM (small-cap tilt).}
\end{itemize}

\subsection{Train/dev/test splits}
\label{app:splits}
Although we do not train the language models, we use splits to avoid \emph{evaluation leakage} from repeated prompt
and guardrail tuning on the same scenarios. We stratify scenarios by (profile, recipe) and assign them to
\textbf{train} (used only to finalize prompts and guardrail solver settings), \textbf{dev} (a one-time check that
the tuned pipeline behaves similarly on unseen scenarios), and \textbf{test} (the final reported results).

\subsection{Scenario JSONL template}
\begin{center}
\small
\begin{minipage}{0.95\linewidth}
\begin{itemize}
    \item[] \texttt{\{}
    \item[] \quad \texttt{"scenario\_id": "AF-001",}
    \item[] \quad \texttt{"as\_of": "2025-09-30",}
    \item[] \quad \texttt{"profile": "Conservative|Moderate|Aggressive",}
    \item[] \quad \texttt{"caps": \{"sigma\_cap":..., "fee\_cap":..., "hhi\_cap":..., "max\_asset":..., "max\_sector":...\},}
    \item[] \quad \texttt{"asset\_universe": ["SPY","VEA","VWO","VGT","XLE","XLF","XLV",}
    \item[] \quad \quad \texttt{"XLY","XLP","XLI","XLRE","IWM","AGG","LQD","IEF","GLD"],}
    \item[] \quad \texttt{"biasmix": "<bias recipe text>",}
    \item[] \quad \texttt{"seeds": [1],}
    \item[] \quad \texttt{"split": "train|dev|test"}
    \item[] \texttt{\}}
\end{itemize}
\end{minipage}
\end{center}

\section{Additional Results}
\subsection{Test first-pass violation rates (Table~\ref{tab:test-viol})}
\begin{table}[t]
\caption{\textbf{Test first-pass violation rate (Wilson 95\% CI).} Violations are measured on the model's raw proposal $w_0$ before projection; lower values indicate better prompt-level compliance, not guardrail effectiveness.}
\label{tab:test-viol}
\vspace{\baselineskip}
\begin{center}
\small
\begin{tabular}{lllll}
\multicolumn{1}{c}{\bf Model} & \multicolumn{1}{c}{\bf Mode} & \multicolumn{1}{c}{\bf $n$} & \multicolumn{1}{c}{\bf Viol.} & \multicolumn{1}{c}{\bf Violation rate (95\% CI)}
\\ \hline
gemini-2.5-flash & critique & 21 & 13 & 0.619 [0.409,0.792] \\
gemini-2.5-flash & direct & 21 & 18 & 0.857 [0.654,0.950] \\
gemini-2.5-flash & sc & 21 & 10 & 0.476 [0.283,0.676] \\
gpt-5-nano & critique & 21 & 17 & 0.810 [0.600,0.923] \\
gpt-5-nano & direct & 21 & 17 & 0.810 [0.600,0.923] \\
gpt-5-nano & sc & 21 & 13 & 0.619 [0.409,0.792] \\
meta-llama/Llama-3.3-70B-Instruct-Turbo & critique & 21 & 12 & 0.571 [0.365,0.755] \\
meta-llama/Llama-3.3-70B-Instruct-Turbo & direct & 21 & 16 & 0.762 [0.549,0.894] \\
meta-llama/Llama-3.3-70B-Instruct-Turbo & sc & 21 & 11 & 0.524 [0.324,0.717] \\
\end{tabular}
\end{center}
\end{table}

\subsection{Test robustness and success rates (Table~\ref{tab:test-robust})}
\label{app:robustness}
\begin{table}[t]
\caption{\textbf{Test robustness and success rates (Wilson 95\% CI).}
Parse-fail counts a scenario only if all $R{=}3$ retries fail strict JSON parsing.
Final feasibility is the post-projection pass rate ($w^*\in\mathcal{C}$).
End-to-end success requires both parsing and final feasibility.}
\label{tab:test-robust}
\vspace{\baselineskip}
\begin{center}
\small
\begin{tabular}{@{}llcccc@{}}
\textbf{Model} & \textbf{Mode} & \textbf{$n$} & \textbf{Parse fails} & \textbf{Parse-fail (95\% CI)} & \textbf{Final pass (95\% CI)} \\
\hline
gemini-2.5-flash & critique & 21 & 0 & 0.000 [0.000, 0.155] & 1.000 [0.845, 1.000] \\
gemini-2.5-flash & direct & 21 & 0 & 0.000 [0.000, 0.155] & 1.000 [0.845, 1.000] \\
gemini-2.5-flash & sc & 21 & 0 & 0.000 [0.000, 0.155] & 1.000 [0.845, 1.000] \\

gpt-5-nano & critique & 21 & 0 & 0.000 [0.000, 0.155] & 1.000 [0.845, 1.000] \\
gpt-5-nano & direct & 21 & 0 & 0.000 [0.000, 0.155] & 1.000 [0.845, 1.000] \\
gpt-5-nano & sc & 21 & 0 & 0.000 [0.000, 0.155] & 1.000 [0.845, 1.000] \\

Llama-3.3-70B & critique & 21 & 0 & 0.000 [0.000, 0.155] & 1.000 [0.845, 1.000] \\
Llama-3.3-70B & direct & 21 & 0 & 0.000 [0.000, 0.155] & 1.000 [0.845, 1.000] \\
Llama-3.3-70B & sc & 21 & 0 & 0.000 [0.000, 0.155] & 1.000 [0.845, 1.000] \\

\end{tabular}
\end{center}
\end{table}

\subsection{Qualitative bias prompt effectiveness (LLM output examples)}
\label{sec:bias-examples}
Table~\ref{tab:bias-examples} provides one concrete held-out test example per BiasMix recipe, showing that the first-pass draft allocations and the model's explanation reflect the intended bias (e.g., VGT/XLE/GLD/IWM tilts). We show direct prompting for brevity; critique/SC exhibit similar thematic emphasis but are omitted due to space.

\begin{table}[t]
\caption{\textbf{Bias prompt effectiveness on held-out test:} representative first-pass LLM draft allocations and short explanation excerpts for each BiasMix recipe (direct prompting shown for compactness).}
\vspace{\baselineskip}
\label{tab:bias-examples}
\begin{center}
\small
\setlength{\tabcolsep}{3pt}
\renewcommand{\arraystretch}{1.15}
\begin{tabular}{l l p{3.5cm} p{4.5cm}}
\hline \\
\multicolumn{1}{c}{\bf Bias recipe} & \multicolumn{1}{c}{\bf Scenario} & \multicolumn{1}{c}{\bf Draft weights (excerpt)} & \multicolumn{1}{c}{\bf Explanation snippet (verbatim)} \\
\hline \\
Anchor tech & AF-051 (Agg.) &
VGT 45\%, SPY 30\%, VEA 15\%, VWO 10\% &
``strong tech tilt via VGT (45\%), aligning with the `Anchor tech' bias'' \\
\rule{0pt}{3ex}
FOMO energy & AF-004 (Cons.) &
XLE 10\%, AGG 40\%, LQD 20\%, IEF 17\% &
``strategic tilt towards XLE (energy) \dots aligning with the FOMO energy bias'' \\
\rule{0pt}{3ex}
Small-cap hype & AF-061 (Agg.) &
IWM 25\%, SPY 30\%, VEA 15\%, VWO 10\% &
``prioritizes \dots small-cap exposure via IWM, aligning with the `small-cap hype' bias'' \\
\rule{0pt}{3ex}
Gold craze & AF-047 (Mod.) &
GLD 20\%, SPY 20\%, AGG 20\%, VEA 15\% &
``a significant gold tilt (GLD) as requested'' \\
\rule{0pt}{3ex}
EM tilt & AF-060 (Agg.) &
VWO 30\%, SPY 30\%, VEA 20\%, VGT 10\% &
``strong EM tilt via VWO, complemented by VEA'' \\
\rule{0pt}{3ex}
US-only bias & AF-020 (Cons.) &
AGG 60\%, SPY 15\%, LQD 10\%, IEF 10\%, XLP 5\% &
``US equity exposure is limited to SPY \dots aligning with the US-only bias'' \\
\rule{0pt}{3ex}
Default inertia & AF-009 (Cons.) &
AGG 100\% &
``default inertia to favor AGG materially \dots a 100\% allocation to AGG'' \\
\rule{0pt}{3ex}
Fee neglect & AF-066 (Agg.) &
SPY 35\%, VEA 20\%, VWO 15\%, IWM 10\%, AGG 10\% &
``diversified \dots and US small-cap (IWM) \dots respecting the fee neglect bias'' \\
\hline
\end{tabular}
\end{center}
\end{table}

\begin{figure}[t]
\begin{center}
    \includegraphics[width=0.95\linewidth]{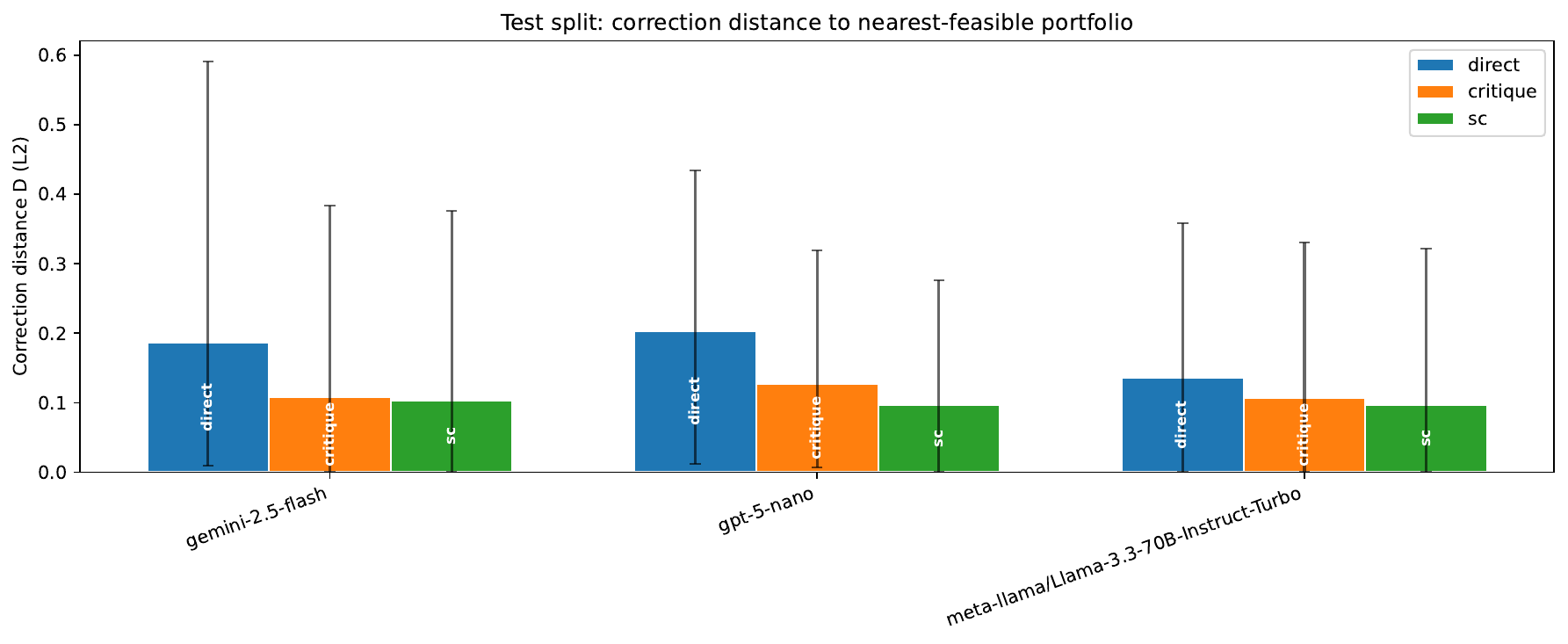}
\end{center}
\caption{Held-out test: correction distance $D=\| w^* - w_0 \|_2$ (profile-pooled) with conservative CI bands.}
\label{fig:D}
\end{figure}

\subsection{Correction Distances Visualization - \ref{fig:D}}
\label{app:correction-distance}

\subsection{Bias analysis on held-out test (violation heatmaps)}
\label{sec:bias-heatmap}
\begin{figure}[t]
\begin{center}
    \includegraphics[width=0.98\linewidth]{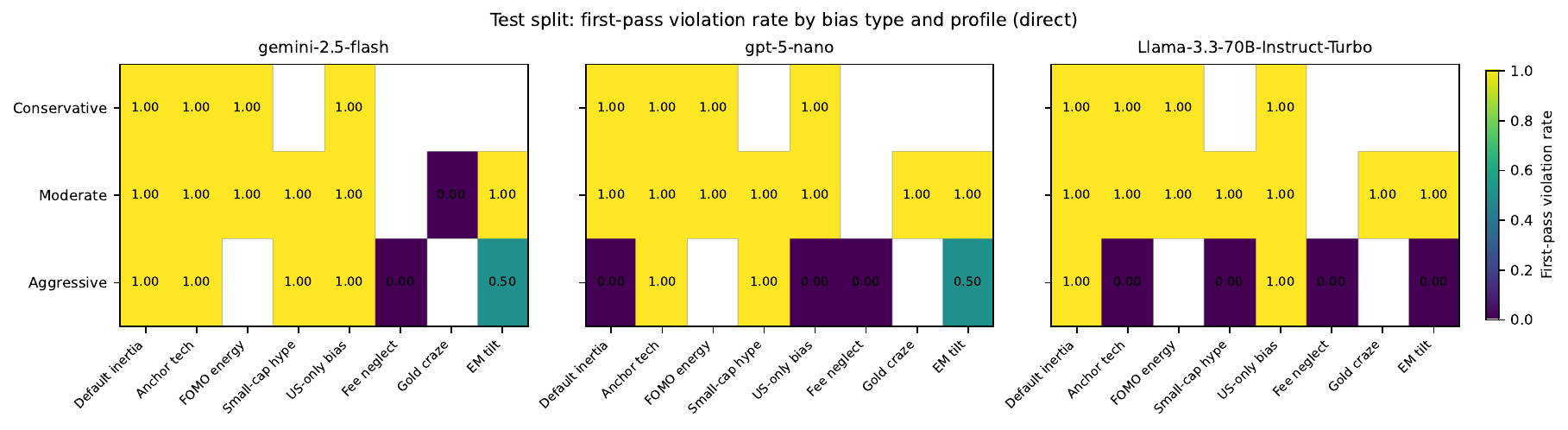}
\end{center}
\caption{\textbf{Test-split bias heatmap (Direct):} first-pass violation rate by profile (rows) and bias type (columns), shown per model (panels).}
\label{fig:bias-heatmap-direct}
\end{figure}

\begin{figure}[t]
\begin{center}
    \includegraphics[width=0.98\linewidth]{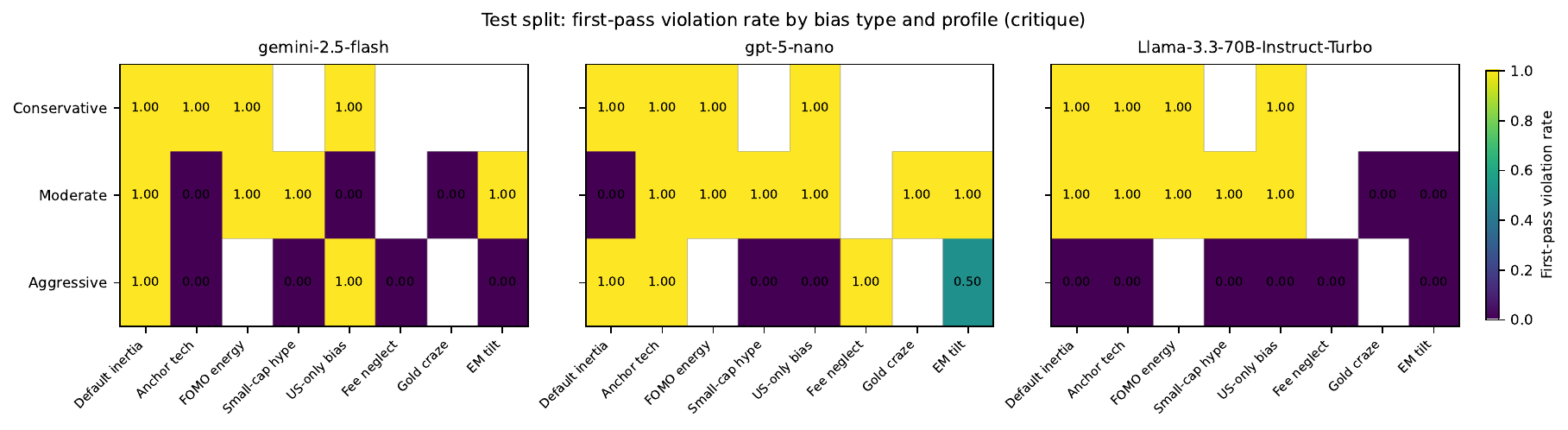}
\end{center}
\caption{\textbf{Test-split bias heatmap (Critique):} first-pass violation rate by profile and bias type, shown per model.}
\label{fig:bias-heatmap-critique}
\end{figure}

\begin{figure}[t]
\begin{center}
    \includegraphics[width=0.98\linewidth]{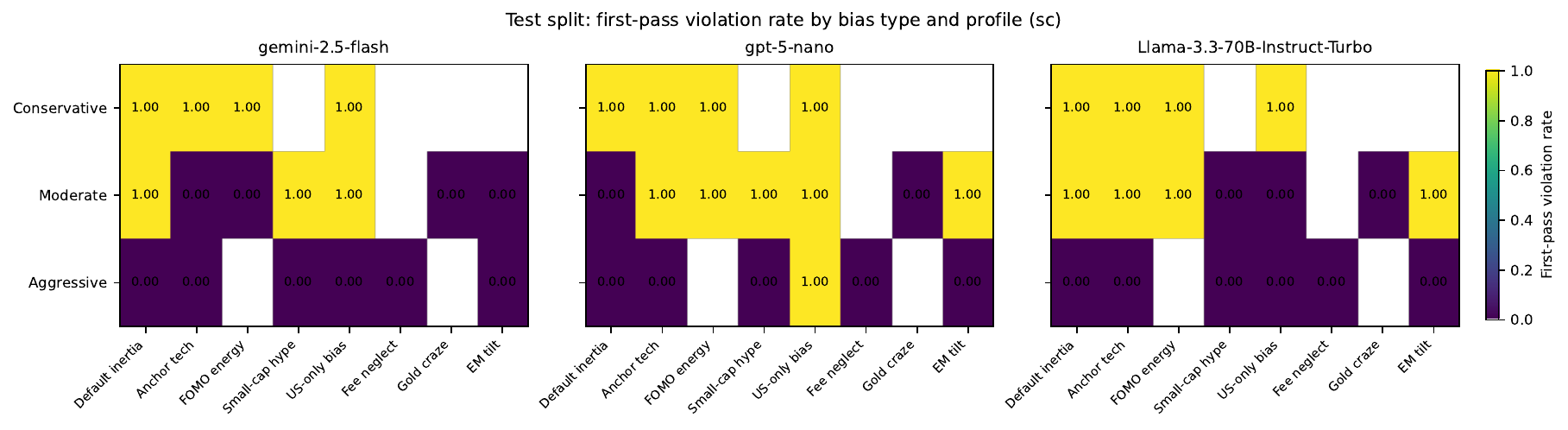}
\end{center}
\caption{\textbf{Test-split bias heatmap (Self-consistency):} first-pass violation rate by profile and bias type, shown per model.}
\label{fig:bias-heatmap-sc}
\end{figure}

To inspect bias-specific failure modes, we compute first-pass violation rates on the \textit{test split} for each (profile, bias type) cell, shown separately by model and prompting mode in Figs.~\ref{fig:bias-heatmap-direct}--\ref{fig:bias-heatmap-sc}.
Cells with no test scenarios for a given (profile, bias) combination are left blank.

\subsection{Sector-level intent preservation (before/after)}
\label{sec:sector-preserve}
\begin{figure}[t]
\begin{center}
    \includegraphics[width=0.98\linewidth]{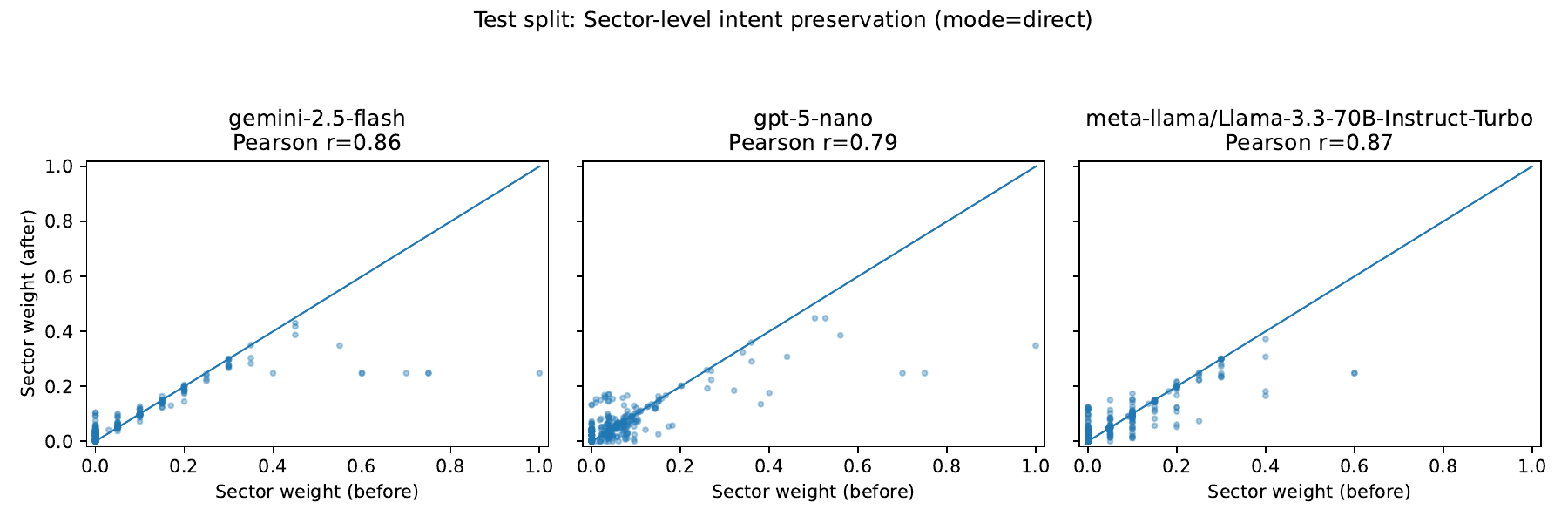}
\end{center}
\caption{\textbf{Sector before/after (Direct, test split):} draft vs.\ projected sector allocations, shown per model (panels).}
\label{fig:sector-preserve-direct}
\end{figure}

\begin{figure}[t]
\begin{center}
    \includegraphics[width=0.98\linewidth]{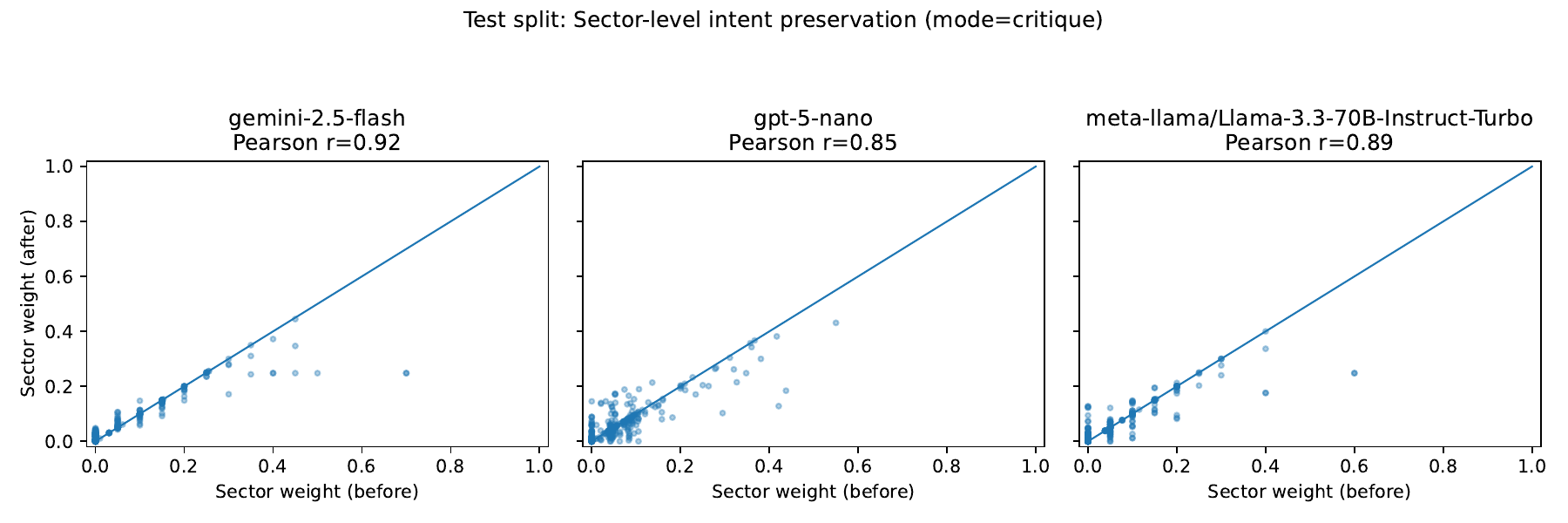}
\end{center}
\caption{\textbf{Sector before/after (Critique, test split):} draft vs.\ projected sector allocations, shown per model.}
\label{fig:sector-preserve-critique}
\end{figure}

\begin{figure}[t]
\begin{center}
    \includegraphics[width=0.98\linewidth]{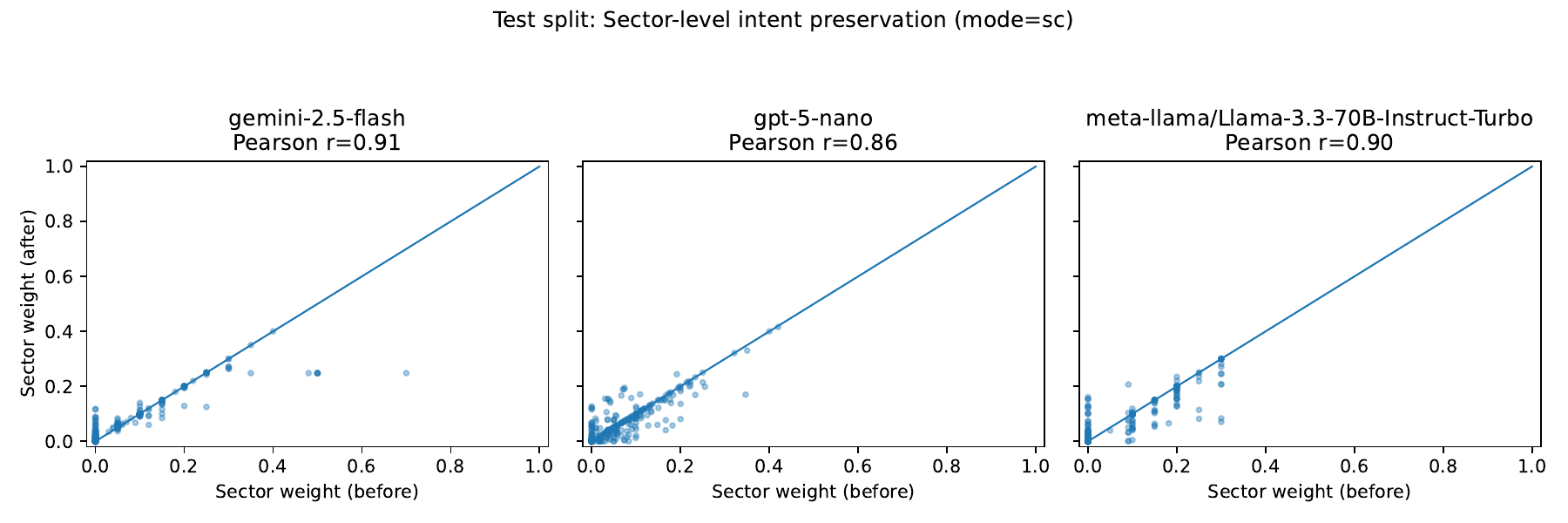}
\end{center}
\caption{\textbf{Sector before/after (Self-consistency, test split):} draft vs.\ projected sector allocations, shown per model.}
\label{fig:sector-preserve-sc}
\end{figure}

To assess intent preservation beyond the L2 correction distance, we compare \emph{sector-sum} allocations before and after projection on the held-out test split.
Each point in Figs.~\ref{fig:sector-preserve-direct}--\ref{fig:sector-preserve-sc} corresponds to a (scenario, sector) pair, plotting the draft sector weight (x-axis) against the post-projection sector weight (y-axis).
Points close to the $y=x$ line indicate that the projection repair preserves the model's sector-level intent while enforcing hard caps.

\subsection{Paired model comparisons (Wilcoxon on correction distance \texorpdfstring{$D$}{D})}
Table~\ref{tab:wilcoxon_all} reports paired Wilcoxon signed-rank tests comparing models on the correction distance $D$
(only scenarios where both models produced a valid run are paired).
We report the Wilcoxon statistic $W$, the unadjusted $p$-value, the effect size $r$, and the Benjamini--Hochberg
FDR-adjusted $p$-value ($p_{\mathrm{FDR}}$). ``Reject'' indicates rejection at FDR $\alpha=0.10$.

\begin{table}[t]
\caption{Paired Wilcoxon tests on $D$ for all model pairs, reported by split and reasoning mode.}
\label{tab:wilcoxon_all}
\vspace{\baselineskip}
\begin{center}
\small
\setlength{\tabcolsep}{3pt}
\begin{tabular}{llcccrcc}
\multicolumn{1}{c}{\bf Split} & \multicolumn{1}{c}{\bf Mode} & \multicolumn{1}{c}{\bf Model A} & \multicolumn{1}{c}{\bf Model B} & \multicolumn{1}{c}{\bf $N$} & \multicolumn{1}{c}{\bf $r$} & \multicolumn{1}{c}{\bf $p_{\mathrm{FDR}}$} 
\\ \hline \\
train & critique & \texttt{gpt-5-nano} & \texttt{meta-llama/Llama-3.3-70B} & 36 & 0.52 & 0.0054 \\
train & critique & \texttt{gemini-2.5-flash} & \texttt{gpt-5-nano} & 36 & 0.30 & 0.0707 \\
train & critique & \texttt{gemini-2.5-flash} & \texttt{meta-llama/Llama-3.3-70B} & 36 & 0.43 & 0.0162 \\
train & direct & \texttt{gpt-5-nano} & \texttt{meta-llama/Llama-3.3-70B} & 36 & 0.47 & 0.0122 \\
train & direct & \texttt{gemini-2.5-flash} & \texttt{gpt-5-nano} & 36 & 0.23 & 0.1686 \\
train & direct & \texttt{gemini-2.5-flash} & \texttt{meta-llama/Llama-3.3-70B} & 36 & 0.35 & 0.0581 \\
train & sc & \texttt{gpt-5-nano} & \texttt{meta-llama/Llama-3.3-70B} & 36 & 0.21 & 0.2021 \\
train & sc & \texttt{gemini-2.5-flash} & \texttt{gpt-5-nano} & 36 & 0.27 & 0.1463 \\
train & sc & \texttt{gemini-2.5-flash} & \texttt{meta-llama/Llama-3.3-70B} & 36 & 0.29 & 0.1463 \\
dev & critique & \texttt{gpt-5-nano} & \texttt{meta-llama/Llama-3.3-70B} & 15 & 0.56 & 0.0417 \\
dev & critique & \texttt{gemini-2.5-flash} & \texttt{gpt-5-nano} & 15 & 0.39 & 0.1186 \\
dev & critique & \texttt{gemini-2.5-flash} & \texttt{meta-llama/Llama-3.3-70B} & 15 & 0.58 & 0.0417 \\
dev & direct & \texttt{gpt-5-nano} & \texttt{meta-llama/Llama-3.3-70B} & 15 & 0.68 & 0.0352 \\
dev & direct & \texttt{gemini-2.5-flash} & \texttt{gpt-5-nano} & 15 & 0.36 & 0.1362 \\
dev & direct & \texttt{gemini-2.5-flash} & \texttt{meta-llama/Llama-3.3-70B} & 15 & 0.54 & 0.0512 \\
dev & sc & \texttt{gpt-5-nano} & \texttt{meta-llama/Llama-3.3-70B} & 15 & 0.33 & 0.1987 \\
dev & sc & \texttt{gemini-2.5-flash} & \texttt{gpt-5-nano} & 15 & 0.46 & 0.1137 \\
dev & sc & \texttt{gemini-2.5-flash} & \texttt{meta-llama/Llama-3.3-70B} & 15 & 0.44 & 0.1314 \\
test & critique & \texttt{gpt-5-nano} & \texttt{meta-llama/Llama-3.3-70B} & 21 & 0.62 & 0.0098 \\
test & critique & \texttt{gemini-2.5-flash} & \texttt{gpt-5-nano} & 21 & 0.36 & 0.1168 \\
test & critique & \texttt{gemini-2.5-flash} & \texttt{meta-llama/Llama-3.3-70B} & 21 & 0.61 & 0.0098 \\
test & direct & \texttt{gpt-5-nano} & \texttt{meta-llama/Llama-3.3-70B} & 21 & 0.54 & 0.0292 \\
test & direct & \texttt{gemini-2.5-flash} & \texttt{gpt-5-nano} & 21 & 0.21 & 0.3491 \\
test & direct & \texttt{gemini-2.5-flash} & \texttt{meta-llama/Llama-3.3-70B} & 21 & 0.42 & 0.0669 \\
test & sc & \texttt{gpt-5-nano} & \texttt{meta-llama/Llama-3.3-70B} & 21 & 0.33 & 0.2055 \\
test & sc & \texttt{gemini-2.5-flash} & \texttt{gpt-5-nano} & 21 & 0.28 & 0.2055 \\
test & sc & \texttt{gemini-2.5-flash} & \texttt{meta-llama/Llama-3.3-70B} & 21 & 0.30 & 0.2055 \\
\hline
\end{tabular}
\end{center}
\end{table}

\noindent\textbf{Interpretation}
Across splits, paired Wilcoxon tests on correction distance $D$ show that \emph{direct} and \emph{critique} modes often yield statistically distinguishable $D$ distributions between certain model pairs after BH-FDR correction (notably comparisons involving \texttt{meta-llama/Llama-3.3-70B-Instruct-Turbo}). In contrast, \emph{sc} mode exhibits fewer significant differences across model pairs (most $p_{\mathrm{FDR}} \ge 0.10$), suggesting that the self-consistency selection tends to reduce cross-model variability in the magnitude of post-generation corrections. We report these as empirical outcomes rather than enforcing an a priori ordering between modes.

\section{Solver Projection Implementation}
\label{app:solver}
\subsection{Solver Settings}
\noindent\textbf{Convexity}
Because \(\Sigma\succeq 0\) and the constraints in \(\mathcal{C}(\theta)\) are convex (simplex, box, linear sector sums,
and convex quadratic bounds on volatility and HHI), Eq.~\eqref{eq:projection} is a convex QCQP.
We implement it in CVXPY with a PSD ``jitter'' \(\Sigma \leftarrow \Sigma + 10^{-10}I\) for numerical stability.
In practice, the quadratic volatility/HHI caps induce a conic form, so we solve the QCQP with SCS \citep{odonoghue2016conic}
(\texttt{eps=1e-5}, \texttt{max\_iters=30000}).
For robustness, we optionally attempt OSQP when CVXPY canonicalizes a given instance to a pure QP (no quadratic caps),
using \texttt{eps\_abs=eps\_rel=1e-7}, \texttt{max\_iter=200000}, and \texttt{polish=True}; otherwise (or if a solver returns a non-optimal status) we fall back to SCS.
This retry logic improves robustness across models and scenarios.

\noindent\textbf{Robustness to solver settings}
Appendix~\ref{app:solver-ablation} reports a small sensitivity study over solver tolerances and iteration limits; feasibility and correction distance remain stable.

\noindent\textbf{Interior margins and ``polish''}
To reduce borderline numerical violations, we solve with small interior margins (e.g., \(\tau=2\times 10^{-3}\) for
asset/sector caps, and \(hhi\_eps=3\times 10^{-4}\) for HHI).
If the resulting solution is still slightly above the HHI cap due to rounding, we re-solve once with a tighter HHI bound
and warm-start from the previous solution. As a final deterministic safeguard, we apply a tiny weight transfer
(from the largest weight to a low-fee diversifier) if HHI remains marginally above the cap.

\subsection{Solver parameter ablation and robustness}
\label{app:solver-ablation}

Table~\ref{tab:solver-ablation} varies numerical tolerances, iteration budgets.
We report (i) solver success rate, (ii) final feasibility after projection, (iii) median correction distance $D=\|w^*-w_0\|_2$, and (iv) mean runtime.
Across settings, the QCQP projection remains robust: SCS \citep{odonoghue2016conic} reliably solves the conic form induced by quadratic caps, while OSQP \citep{stellato2020osqp} primarily serves as an optional fast-path when an instance reduces to a pure QP.

\begin{table}[t]
\caption{Solver robustness ablation across different settings.}
\label{tab:solver-ablation}
\begin{center}
\small
\setlength{\tabcolsep}{3pt} 
\begin{tabular}{lrrrrrr}
\hline
\multicolumn{1}{c}{\bf Setting} & \multicolumn{1}{c}{\bf Solve} & \multicolumn{1}{c}{\bf Final} & \multicolumn{1}{c}{\bf Solved by} & \multicolumn{1}{c}{\bf Median} & \multicolumn{1}{c}{\bf Mean time} & \multicolumn{1}{c}{\bf N} \\
 & \multicolumn{1}{c}{\bf success \%} & \multicolumn{1}{c}{\bf feasible \%} & \multicolumn{1}{c}{\bf OSQP \%} & \multicolumn{1}{c}{\bf D} & \multicolumn{1}{c}{\bf (ms)} & \\
\hline
Default (paper) & 100.000000 & 100.000000 & 0.000000 & 0.066900 & 61.320000 & 189 \\
SCS tighter tol & 100.000000 & 100.000000 & 0.000000 & 0.066900 & 59.290000 & 189 \\
SCS looser tol & 100.000000 & 100.000000 & 0.000000 & 0.066900 & 59.730000 & 189 \\
More jitter & 100.000000 & 100.000000 & 0.000000 & 0.066900 & 61.180000 & 189 \\
No OSQP fast-path & 100.000000 & 100.000000 & 0.000000 & 0.066900 & 58.160000 & 189 \\
\hline
\end{tabular}
\end{center}
\end{table}

\section{Reproducibility + Parsing/Decoding}
\label{app:reproducibility}
\subsection{Implementation notes}
We log model name, mode, prompt hash, seed, parse failures, and before/after metrics for reproducibility. We also
run covariance and cap sanity checks (e.g., feasibility probes and eigenvalue diagnostics) during dataset creation;
details are included in the appendix.

\subsection{Decoding parameters}
We use a strict JSON-only output contract for all modes.
Temperatures are mode-specific but shared across models: direct $T{=}0.10$, critique $T{=}0.20$, and self-consistency $T{=}0.28$.
When supported, we set $\texttt{top\_p}{=}0.75$ and $\texttt{max\_tokens}{=}1000$.
For GPT-5 nano, temperature/top-$p$ are not exposed in the API used; we keep default sampling settings and set
$\texttt{reasoning\_effort}{=}\texttt{low}$.

\subsection{Retry + parsing logic}
Each query is attempted up to $R{=}3$ times if strict JSON parsing fails.
A run is counted as a \emph{parse failure} only if all $R$ attempts fail.
Parsing enforces a top-level object with exactly two keys: \texttt{weights} (a ticker-to-weight map) and \texttt{explanation} (a short string).
Weights are coerced to nonnegative reals and renormalized to sum to 1. If the model returns a singleton list containing the object, we unwrap it.

\section{Statistical and Experimental Protocol Details}
\label{app:statistical-details}
\subsection{Primary endpoints}
We pre-register two primary endpoints:
(1) \textbf{First-pass violation rate.} For each cap (and the any-cap aggregate), we mark a violation on $w_0$ if the corresponding constraint in $\mathcal{C}$ is not satisfied (e.g., $\sigma(w_0)>\sigma_{\mathrm{cap}}$).; and
(2) \textbf{correction distance} $D=\|w^*-w_0\|_2$ (Euclidean distance) measuring how much the guardrail must change
the model output to satisfy hard caps.

\subsection{Secondary endpoints}
We report before/after deltas in volatility $\sigma$, fee burden (WAER), and concentration (HHI), as well as parse-failure rate and end-to-end success (valid JSON + post-generation feasibility).
\textbf{(i) Parse failure rate.} A scenario is a parse failure if all $R{=}3$ retries fail strict JSON parsing.
\textbf{(iii) Final feasibility rate.} A scenario is finally feasible if $w^*\in\mathcal{C}$ (all caps satisfied after projection).
\textbf{(iv) End-to-end success.} A scenario is end-to-end successful if it parses \emph{and} is finally feasible.

\subsection{Confidence intervals}
For violation rates (proportions), we compute Wilson 95\% confidence intervals \citep{wilson1927}.
For $D$ and metric deltas, we compute bootstrap 95\% confidence intervals by resampling scenarios with replacement
(10{,}000 resamples) \citep{efron1993}.

\subsection{Model comparisons and multiple testing}
To compare models on $D$ under the same scenarios, we use paired Wilcoxon signed-rank tests on per-scenario differences \citep{wilcoxon1945,demsar2006,dror2018hitchhiker}.
With three models, there are three model pairs per mode. Across three modes and three splits,
this yields $3\times 3\times 3=27$ pairwise tests (and 36 including an ``all-splits'' aggregate view).
We control false discovery using Benjamini--Hochberg FDR at $q=0.10$ \citep{benjamini1995} over the family of pairwise tests and report $p_{\mathrm{FDR}}$.

\subsection{Split-wise reporting}
Train is used only to finalize prompts and solver settings. Dev is reassure settings. We do not re-tune based on test outcomes, instead we
report test metrics (with confidence intervals) as our main evidence of generalization across unseen scenarios and paired comparisons as the main empirical results.

\section{Additional discussion: universe design and generalization}
\label{app:scope}

\textbf{Universe design.}
We adopt a fixed 16-ETF universe as an \emph{intentional design choice} for controlled diagnosis.
It is large enough to express meaningful diversification patterns (U.S., ex-U.S., EM, sectors, bonds, gold) while remaining small enough to (i) make parsing failures and constraint-violation patterns interpretable, (ii) keep covariance and sector mappings stable and auditable, and (iii) enable systematic variation across bias prompts and risk profiles without confounding effects from universe size.

\textbf{Generalization.}
The projection-based guardrail is not tied to 16 assets; it extends to larger universes given fees, sector mappings, and a covariance estimate.
Generalization to hundreds of ETFs and time-varying risk models is left to future work, where scaling behavior, universe-dependent concentration effects, and covariance estimation error become central.

\section{Extended related work (moved from main)}
\label{app:relwork}

\paragraph{Guardrails, auditing, and post-hoc verification.}
Safety and controllability toolkits (e.g., NeMo Guardrails \citep{rebedea2023nemoguardrails}), safety classifiers (e.g., Llama Guard \citep{meta2023llamaguard}), and instruction/alignment methods such as RLHF and Constitutional AI \citep{ouyang2022instruct,bai2022constitutional} primarily aim to filter unsafe content or enforce conversational policies.
A complementary line of work emphasizes \emph{auditing} and \emph{verification} of LLM outputs, especially in high-stakes settings: rather than trusting a single generation, systems instrument the generation process with structured outputs, validation checks, and feedback loops \citep{raji2020closing,ribeiro2020checklist}.
Constrained generation methods provide decoding-time guarantees by enforcing lexical or structural constraints during generation \citep{hokamp2017gridbeam}.
In contrast to content safety, our setting requires satisfying quantitative constraints that are naturally expressed as convex restrictions over portfolio weights; this motivates optimization-based post-processing as an auditable repair layer rather than purely text-level controls \citep{boyd2004convex}.

\paragraph{Structured generation and constrained outputs.}
A practical prerequisite for auditing is that model outputs are machine-checkable.
Prior work on structured outputs and constrained/guided generation \citep{hokamp2017gridbeam,willard2023efficientguided} motivates enforcing schemas so downstream validators can reliably parse and evaluate outputs.
Our pipeline adopts this principle by requiring a strict JSON schema for allocations, enabling deterministic constraint checking and repair.

\paragraph{Constrained portfolio optimization.}
Classical mean--variance optimization \citep{markowitz1952} and later robust/regularized variants \citep{jagannathan2003nogo} motivate using covariance structure and explicit constraints in portfolio construction.
Our work uses these tools as a \emph{repair operator} applied to an LLM draft, aligning with the broader view that LLM outputs may require deterministic verification in high-stakes domains.
Unlike traditional optimization, we do not assume the portfolio is being constructed from first principles to maximize an objective; instead, we minimize deviation from the model's proposal while enforcing KYC-style caps.

\paragraph{Reasoning modes and sampling.}
Prompting strategies such as chain-of-thought \citep{wei2022cot}, self-consistency \citep{wang2022selfconsistency}, and self-refinement/critique \citep{madaan2023selfrefine} can improve reasoning reliability but remain stochastic.
We therefore treat mode comparisons as empirical outcomes and quantify uncertainty with confidence intervals and paired tests, and we position deterministic post-generation verification/repair as the mechanism that guarantees compliance regardless of prompting mode.

\end{document}